\documentclass[journal]{IEEEtran}

\usepackage{amssymb}
\usepackage{optidef}
\usepackage{algorithm}
\usepackage{algorithmic}
\usepackage[table,xcdraw]{xcolor}

\newcommand{\mynorm}[1]{\left \| #1 \right \|}

\usepackage{hyperref}

\begin{document}
%
\title{Adaptive Clustering of Robust Semantic Representations for Adversarial Image Purification}


\author{\IEEEauthorblockN{Samuel Henrique Silva,
Arun Das,
Ian Scarff, and
Peyman Najafirad}\\
\IEEEauthorblockA{Secure AI \& Autonomy Laboratory}
\thanks{This work has been submitted to the IEEE for possible publication. Copyright may be transferred without notice, after which this version may no longer be accessible.}
\thanks{The authors gratefully acknowledge the use of the services of Jetstream cloud.}
\thanks{S. H. Silva, A. Das, I. Scarff, and P. Najafirad are members of the Secure AI and Autonomy Laboratory, University of Texas at San Antonio, San Antonio, TX, 78249 USA. e-mail: (samuelhenrique.silva, arun.das, ian.scarff, peyman.najafirad)@utsa.edu. \textit{Corresponding author: Peyman Najafirad}.}
\thanks{P. Najafirad is also with the Department of Information Systems and Cyber Security, University of Texas at San Antonio, San Antonio, TX, 78249 USA.}
}

%

\IEEEtitleabstractindextext{%
\begin{abstract}
Deep Learning models are highly susceptible to adversarial manipulations that can lead to catastrophic consequences. One of the most effective methods to defend against such disturbances is adversarial training but at the cost of generalization of unseen attacks and transferability across models. In this paper, we propose a robust defense against adversarial attacks, which is model agnostic and generalizable to unseen adversaries. Initially, with a baseline model, we extract the latent representations for each class and adaptively cluster the latent representations that share a semantic similarity. We obtain the distributions for the clustered latent representations and from their originating images, we learn semantic reconstruction dictionaries (SRD). We adversarially train a new model constraining the latent space representation to minimize the distance between the adversarial latent representation and the true cluster distribution. To purify the image, we decompose the input into low and high-frequency components. The high-frequency component is reconstructed based on the most adequate SRD from the clean dataset. In order to evaluate the most adequate SRD we rely on the distance between robust latent representations and semantic cluster distributions. The output is a purified image with no perturbation. Image purification on CIFAR-10 and ImageNet-10 using our proposed method improved the accuracy by more than 10\% compared to state of the art results.
\end{abstract}

}

\maketitle

\IEEEdisplaynontitleabstractindextext

%
\IEEEpeerreviewmaketitle

\section{Introduction}
\label{introduction}


\IEEEPARstart{A}{s a rapidly} developing area of research, Deep Learning (DL) \cite{lecun2015deep} represented a change in the way we interpret and make decisions from data. Since the increase in popularity of DL methods, greatly supported by cloud environments \cite{stewart2015jetstream, keahey2019chameleon}, the vanilla DL models and its variations have been applied in many scientific breakthroughs. From predicting DNA enhancers \cite{yang2017biren}, disease prediction (\cite{das2018distributed}), to natural language processing\cite{ebadi2021memory}, speech recognition \cite{amodei2016deep}, and robotics \cite{silva2020temporal, silva2019cooperative}. DL models are known to perform well in an independent and identically distributed (\textit{i.i.d.}) setting, when test and train data are sampled from the same distribution \cite{schneider2020improving}. However, in many real applications the \textit{i.i.d.} assumption does not hold \cite{chacon2019deep}. Furthermore, the shift in distribution can be artificially introduced by adversarial attacks. Initially exposed by \cite{szegedy2013intriguing}, adversarial attacks are small additive perturbations that, when combined with the input data, cause models to generate wrong predictions with high confidence. The effects of these perturbations in DL models can generate catastrophic consequences in safety critical applications. Additionally, adversarial attacks are mostly imperceptible to the human eye, can be easily transferable across models, and depend on relatively little information about the target model, as seen in \cite{chen2020hopskipjumpattack, ru2020bayesopt}. Hardly distinguishable from clean images, adversarial examples largely lie in the tail of the dataset distribution used in the model training \cite{song2018pixeldefend}. 
\begin{figure}
    \centering
    \includegraphics[width=\linewidth]{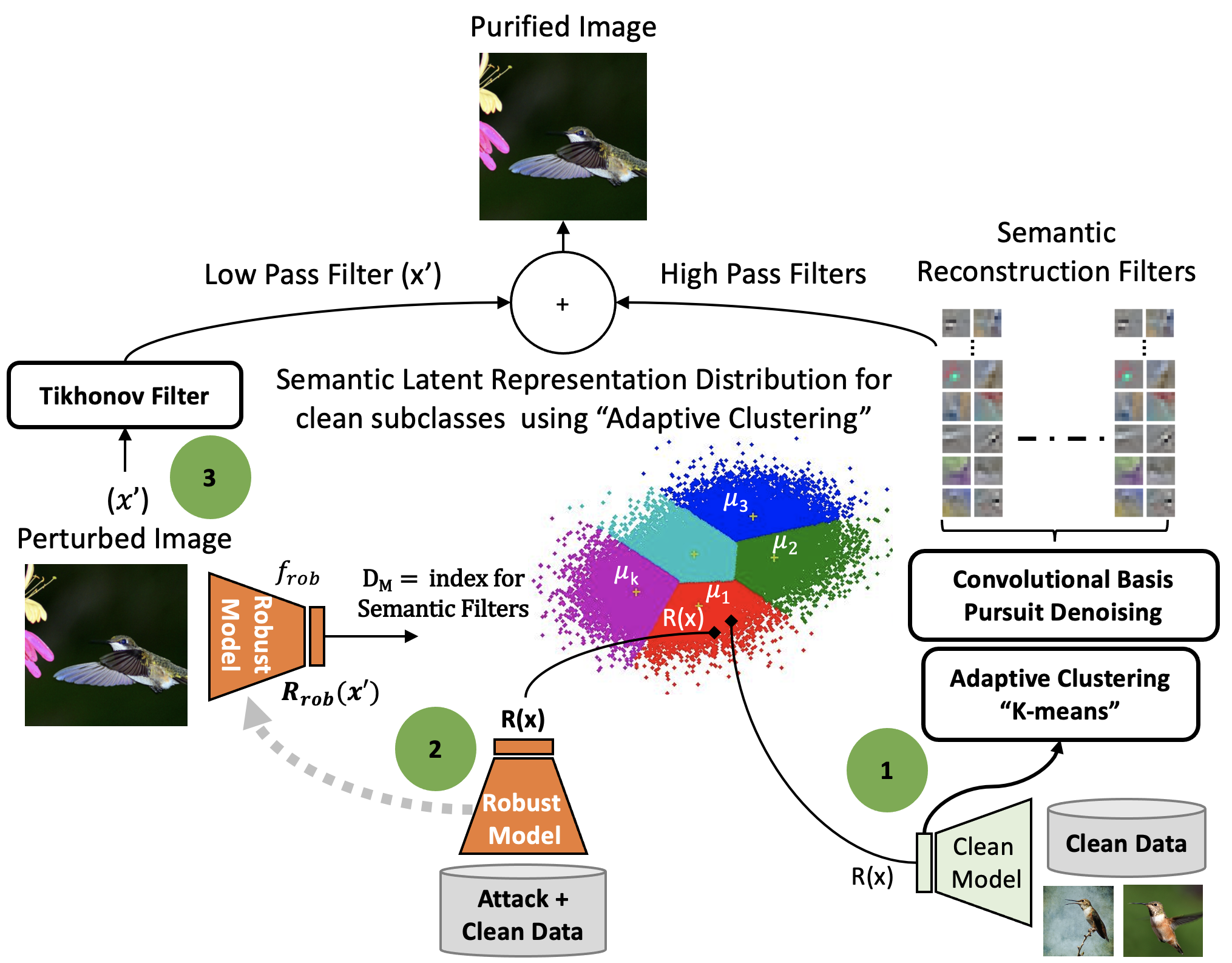}
    \caption{We use both adversarial training and input transformation through convolutional sparse coding to extract robust semantic features and purify adversarial images.}
    \label{fig:general_architecture}
\end{figure}

The methods and effects of adversarial attacks are thoroughly detailed in the literature \cite{silva2020opportunities}. In a similar, way techniques to build robust models against adversarial attacks, either through adversarial training (AT) or input transformation (IT) have received great attention by researchers recently \cite{Madry2018, wong2019fast, guo2018countering}. Whereas, the existence of such small perturbations capable of changing the model's output associated to the constant emergence of new attack methods indicates that current supervised methods and defenses fail to identify true robust features that establish a better causal relation between input and labels. Additionally, the lack of generalization of AT models to unseen attacks indicates that these defenses memorize the attack pattern or pick representations that may not present the desired robustness against small perturbations. To overcome this problem, we go beyond standard AT and combine it with IT to learn semantic representations that demonstrate robustness against small input variations, capture the semantic correlations of the clean dataset, and purifies the adversarial images by removing the adversarial perturbations. We address these tasks in a 3-stage algorithm that achieves performance on par with state-of-the-art (SOTA) AT methods, generalizes to unseen attacks, and is task and model agnostic. 

In our algorithm, represented in \autoref{fig:general_architecture}, we initially train a baseline model with clean images. The baseline model is used to extract the latent representations for all images in the training dataset. Even within the corresponding classes, the latent representations present high variability that induces the model to learn complex and susceptible to manipulation representations. We adaptively cluster the latent representations, so that we create clusters of semantically similar representations. Moreover, we extract the distribution of the latent representations, and from the originating images, we learn semantic reconstruction dictionaries. On the second stage, we train a semantic robust model by modifying the standard adversarial training. We constrain the latent representations to minimize the distance between adversarial representations and the distribution of the clean cluster to which that sample should belong. By constraining the latent representations, we enable the model to extract similar features for clean and adversarial samples. Finally, in the third stage, we purify the adversarial images by reconstructing the high-frequency components of this image using the reconstruction dictionaries obtained from the clean samples in stage one.

More specifically our paper has the following contributions:
\begin{itemize}
    \item We propose a new adversarial training schema, minimizing the distance between the distribution of features extracted from the clean dataset and features from adversarial inputs, improving the generalization to multiple unseen attacks.
    \item We propose an adaptive clustering algorithm for robust semantic representation of latent space to generate multiple reconstruction dictionaries within a class of objects based on their feature similarities and dissimilarities.
    \item We extensively evaluate qualitatively and quantitatively the efficiency of our proposed method in challenging datasets such as CIFAR-10 and ImageNet-10.
\end{itemize}

\section{Related Work}
\label{gen_inst}

\begin{figure*}
    \centering
    \includegraphics[width=\linewidth]{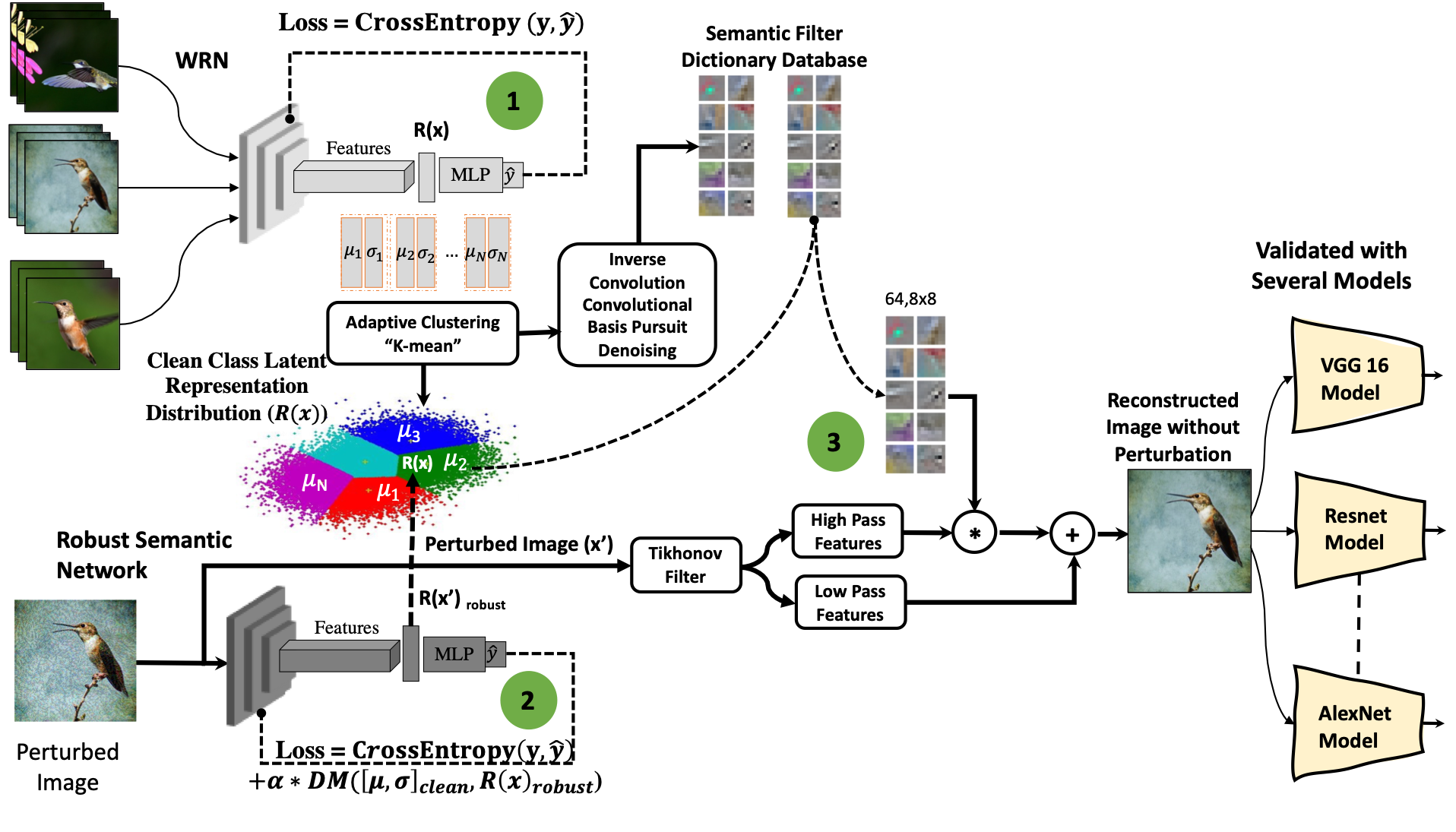}
    \caption{Our purification method combines adversarial training with input transformation. We initially train $f_{bsl}$ in the clean dataset. Based on $f_{bsl}$ we extract clean reference distributions to train our robust model, $f_{rob}$. We cluster the clean input images based on their latent representations and train semantic reconstruction dictionaries based on these clusters. We purify the adversarial inputs with the dictionary from a cluster that minimizes the distance from extracted latent representation and cluster distributions.}

    \label{fig:overall_architecture}
\end{figure*}

\textbf{Adversarial Attacks:} Since the first introduction of adversarial attacks in \cite{szegedy2013intriguing, biggio2013evasion}, the field has seen a great number of contributions from the community \cite{goswami2018unravelling, jia2019fooling, cao2019adversarial, xiao2018spatially}. Adversarial attacks are small, norm constrained perturbations injected in the test data at inference time that are capable of fooling a target model. The generation of such perturbations can be both in a white box setting \cite{Papernot2016, Carlini2018, moosavi2017universal, morgulis2019fooling}, or black box setting \cite{morgulis2019fooling, chen2020hopskipjumpattack, gleave2019adversarial}. With respect to white-box attacks, we assume that the attacker knows all the information related to the data, model architecture, model parameters, gradient, and any other information related to the model. Such attacks are the most efficient in terms of misclassification rate, but these cases are mostly impractical in real-world given the required amount of information concerning the target model. In most cases, attackers do not have access to the model, mostly its outputs. Alternatively, at a small cost to the misclassification rate, black box attacks require only few queries to the model's output to generate perturbations capable of fooling the target model \cite{simonyan2014very, he2016deep, zagoruyko2016wide}. We classify our image purification experimental setting as grey-box. We assume the attacker has full knowledge of the target model, but cannot extract gradient information from our transformation, due to the nature of the Convolutional Basis Pursuit Denoising algorithm. Our Robust Semantic Training is trained and tested in a white-box setting. More specifically we evaluate our defense capabilities against FGSM \cite{goodfellow2015explaining}, BIM, DeepFool \cite{moosavi2016deepfool}, and CW \cite{carlini2017towards} attacks. Our robust model is trained with PGD \cite{Madry2018}.

\textbf{Adversarial Training:} Adversarial training techniques modify the optimization objective of the objective model by incorporating an empirical maximum loss term in the objective, turning the training process into a min-max optimization. Several publications have been presented in which the empirical method to calculate the loss was the main contribution \cite{goodfellow2015explaining, Madry2018, Kannan2018}. Moreover, variations on the feature mapping to improve the accuracy of (re)trained models were proposed by \cite{song2019robust}, and even a tighter upper bound for the max in the loss was proposed by \cite{Zhang2019}. While promising and shown to improve robustness even for large models (ImageNet \cite{Xie2019}), these models come with a drawback which, when instantiated in practice with the approximation heuristics, they are unable to provide robustness guarantees, certifications, or even generalize to unseen attacks. This class of defenses, even though very practical to implement, are model and attack dependent.

\textbf{Input Transformation Defenses:}  Input transformation methods propose to defend against adversarial attacks by composing a set of transformations to adversarial images, such that the output has minimum influence of the adversarial perturbation. \cite{das2018shield} used JPEG compression as a countermeasure to the pixel displacement generated by the adversarial attacks. \cite{guo2018countering} proposed a combination of total variation minimization and image quilting to defend against adversarial attacks. Even though these transformations are nonlinear, a neural network was used to approximate the transformations, making them differentiable, and consequently easier to obtain the gradient. \cite{raff2019barrage} proposed an ensemble of weak transformation defenses to improve the robustness of the models. Among the transformations included in the defense are color precision reduction, JPEG noise, Swirl, Noise Injection, FFT perturbation, Zoom Group, Color Space Group, Contrast Group, Grey Scale Group, and Denoising Group. Input transformation methods are in nature model independent, which improves transferability but limits efficiency as in general model is unaware of the important input features which need restoration to improve target model's accuracy.


\section{Methodology}

\subsection{Method Overview}
\label{sec:method_overview}

Given a model class $\mathcal{F}$ and a dataset $\mathcal{D} = \{x_i, y_i\}_{i=1}^n$ of feature vectors $x_i \in \mathcal{X} \subseteq \mathbb{R}^m$, where $\mathcal{X}$ is the feature space, and labels $y_i$ from some label set $\mathcal{Y}$, assuming up to $c$ possible values to $y_i$, corresponding to the class set $\mathcal{C}$. This dataset is typically assumed to be sampled \textit{i.i.d} from the distribution, $\mathcal{P}$, which is unknown. A classification model is a probability estimator, which maximizes the true class probability, $p(y_i|x_i)$. On the contrary, an adversarial perturbation is an additive noise $\delta$, which maximizes another class probability such that the model's decision changes:
\begin{equation*}
    d(x_i, x_i+\delta)\leq \epsilon,~ and~ f(x) \neq f(x') 
\end{equation*}
in which $\epsilon$ is the maximum allowed perturbation, and $d(.,.)$ is some specified distance. Our proposed method purifies adversarial inputs $x_i' = x_i+\delta$, such that $f(T(x+\delta)) = f(T(x)) = f(x)$. Our method is a composition of adversarial training and input transformation, and is divided in 3 stages: \textit{i)} - Baseline Training; \textit{ii)} - Robust Semantic Training; and \textit{iii)} - Robust Semantic Feature Purification. Algorithm \ref{alg:overall_algorithm}, describes our approach. All details and notations are presented in the following sections of this paper. \autoref{fig:overall_architecture} represents all stages of our algorithm in detail.

In \textit{i)} - Baseline Training, we train a Wide-Resnet model \cite{zagoruyko2016wide} that achieves high classification accuracy. We use this model to extract all the latent representations $R(x) \in \mathbb{R}^k$. For each class, we adaptively cluster all $R(x)$ that share semantic similarities, generating a cluster set $\Psi^c = \{\Psi^c_1, \Psi^c_2, ...\}$, with an optimal number of clusters for class $c$. Based on the cluster of $R(x)$ we cluster the originating inputs $x$, keeping track of images and clusters associations. For each $\Psi^c_i$, we calculate the parameters defining their distribution, $(\mu^c_i, \sigma^c_i)$. In addition, for each cluster, based on the originating images $x$, we generate a sparse code reconstruction dictionary, using Convolutional Dictionary Learning (CDL). The details on the adaptive clustering, distribution calculation, and dictionary learning are discussed in the following sub-sections. 

In \textit{ii)} - Robust Semantic Training, we train a robust model $f_{rob}$, which extracts robust latent representations which lie on the same distribution independently of weather $x_i$ is an adversarial input or a clean input. To achieve such similarity, we constrain the latent representation to minimize the distance between adversarial latent representation and clean sample cluster distribution. The robust model can robustly classify the input, but more importantly, generate a latent representation that is close to the one that would be generated by a clean sample. 

In \textit{iii)} - Robust Semantic Feature Purification, we purify $x_i$ by reconstructing the high frequency components of $x_i$ with its semantic reconstruction dictionary learned in \textit{(i)}. We use $f_{rob}$ to extract robust latent representations from $x_i$, and match them with the semantic reconstruction dictionary that minimizes the distance between $R(x_i)$ and $C_i$ obtained in \textit{(i)}. Following the semantic dictionary matching, we decompose $x_i$ into its low and high-frequency components. We use convolutional sparse coding to reconstruct the high-frequency components of $x_i$ and combine with the low-frequency to generate the purified and transformed version of $x_i$, $T(x_i)$. 

\subsection{Problem Definition}
\begin{algorithm}[!b]
   \caption{Adversarial Image Purification}
   \label{alg:overall_algorithm}
\begin{algorithmic}
   \STATE {\bfseries Input:} $\mathcal{D}_{tr}$
   \vspace{1mm}
   \STATE {\bfseries \textit{Stage i} - Baseline Training}
   \vspace{1mm}
   \STATE $f_{bsl} \leftarrow$ \textit{standard-training($\mathcal{D}_{tr}$)}
   \STATE $\mathcal{R} \leftarrow$ \textit{extract-latent-representation($f_{bsl}$, $\mathcal{D}_{tr}$)}
   \FOR{$c \in \mathcal{C}=\{C_1,C_2, ...\}$ } 
   \STATE $\psi_c$ = \textit{elbow($R(x_i)$)}, $\forall x_i \in c$
   \STATE $\Psi_c$ = \textit{k-means($R(x_i)$, $\psi_c$)}, $\forall x_i \in c$
   \FOR{$j \leq \psi_c$}
   \STATE $\mu_c^j = \frac{1}{size(\Psi_c^j)} \sum x_i, ~ \forall x_i \in \Psi_c^j$
   \STATE $\sigma_c^j = \mathbb{E}[(\Psi_c^j - \mathbb{E}[\Psi_c^j])(\Psi_c^j - \mathbb{E}[\Psi_c^j])^T], ~ \forall x_i \in \Psi_c^j$
   \STATE $\Phi_c^j = $\textit{concat(CDL($x_i$), ($x_i$), ($\mu_c^j, \sigma_c^j$))}$, ~ \forall x_i \in \Psi_c^j$
   \ENDFOR
   \ENDFOR
   \vspace{1mm}
   \STATE {\bfseries \textit{Stage ii} - Robust Semantic Training}
   \vspace{1mm}
   \FOR {$(x,y) \in \Phi$}
   \STATE $x' = $ \textit{adversarial($x$)}
   \STATE $\hat{y}, R(x') \leftarrow f_{rob}(x')$
   \STATE $l = loss(\hat{y},y) + \lambda*dist(R(x'), (\mu, \sigma))$
   \STATE $f_{rob} \leftarrow update(f_{rob}, l)$
   \ENDFOR
   \vspace{1mm}
   \STATE {\bfseries \textit{Stage iii} -  Robust Semantic Feature Purification}
   \vspace{1mm}
   \STATE $x_{low}, x_{high}$ = \textit{tikhonov}($x_i$)
   \STATE $R(x_i)$ = $f_{rob}(x_i)$
   \STATE $\Phi_{rec}$ = $argmin_\Phi ~dist(\Phi, R(x_i))$ 
   \STATE $x^{rec}_{high}$ = \textit{CBPDN($x_{high}$, $\Phi_{rec}$)}
   \STATE $x_{pur}$ = $x^{rec}_{high} + x_{low}$
\end{algorithmic}
\end{algorithm}
The objective of supervised methods is to find a model $f \in \mathcal{F}$, such that:
\begin{equation*}
    \mathbb{E}_{(x,y) \sim \mathcal{P}}[l(f(x), y)] \leq \mathbb{E}_{(x,y)\sim \mathcal{P}}[l({f}'(x), y)] ~ \forall ~ {f}' \in \mathcal{F},
\end{equation*}
where the loss function, $l(f(x),y)$, measures the error that $f(x)$ makes in predicting the true label $y$. In practice $\mathcal{P}$ is unknown, and in replacement we use a training data $\mathcal{D}_{tr}$, in order to find a candidate $f$ which is a good approximation of the labels actually observed in this data. Which raises the problem, known as \textit{empirical risk minimization} (ERM):
\begin{mini}|s|
{\theta}{\sum_{i\in\mathcal{D}_{tr}} l(f(x_i;\theta), y_i) + \lambda \rho(\theta),}
{\label{eq:supervised_learning}}{ }
\end{mini}
in which $\theta$ are the model parameters and $\rho(\theta)$ is a regularization function to constrain the changes of the model parameters at each learning step. We refer to \autoref{eq:supervised_learning} as the baseline model training.
The standard premise in DL is that finding the optimum parameters in \autoref{eq:supervised_learning} would deliver high performance on unseen data draw from the same distribution. On the contrary, the baseline objective from (\ref{eq:supervised_learning}) is highly vulnerable to small perturbations, crafted by adversarial algorithms. In a general formulation, these perturbations are generated such that:
\begin{maxi}|s|
{\mynorm{\delta}_2 \leq \epsilon}{l(f(x_i + \delta;\theta), y_i). }
{\label{eq:adversarial_attack}}{ }
\end{maxi}
By introducing the perturbation $\delta$ in the test data $\mathcal{D}_{te}$, the actual test distribution is shifted to the tail of the training distribution, affecting the performance of $f(x)$ when evaluated in $\mathcal{D}_{te}$. A natural approach to mitigate the effects of these manipulations is the introduction of the adversarial examples in \autoref{eq:supervised_learning}, known as the \textit{robust optimization framework}: 
\begin{equation}
    \min_\theta ~ \mathbb{E}_{(x,y) \sim \mathcal{D}} \max_{\delta \in \Delta}~ l(f(x+\delta), y) +  \lambda \rho(\theta),
\label{eq:robust_optimization}
\end{equation}
which incorporates the adversarial samples in the training data. \autoref{eq:robust_optimization} is the \textit{standard adversarial training}. The standard adversarial training addresses the immediate issue of samples generated by the technique used to empirically approximate \autoref{eq:adversarial_attack} for model $f(.)$. It has been shown by many publications \cite{silva2020opportunities}, that this formulation does not generalize to unseen attacks. Moreover, the need to retrain every model can make such formulation very expensive.

In the baseline model (\ref{eq:supervised_learning}) and adversarial model (\ref{eq:robust_optimization}), the convolutional layers learn to extract latent representations $R(x) \in \mathbb{R}^k$ which are meaningful to the Fully Connected layers of the model. $R(x)$ is the output of the last layer before the fully connected layers of the model. As shown in \cite{engstrom2019adversarial}, considerably different inputs can generate fairly similar latent representations. This evidence shows that even though latent representations are relevant condensed features for the model's fully connected layers, the similarity between latent representations of different classes is the source of the model's susceptibility to adversarial attacks.

\subsection{Baseline Training}
\label{baseline_training}

In the baseline training, we construct all the references for robust model training and image purification. As mentioned in \autoref{sec:method_overview}, we train a Wide Residual Network (WRN) for a classification task. The WRN model has been used in several publications as a benchmark for adversarial training in classification models and its feature extraction capability. We refer to the baseline model $f_{bsl}$, a model trained without any adversarial training or robustness technique, except standard ones such as: Batch Normalization, Dropout, and Parameter Regularization.

The baseline model $f_{bsl}$ is trained on $\mathcal{D}_{tr}$. It is required that $f_{bsl}$ model with good accuracy the distribution of $\mathcal{D}_{tr}$, and consequently achieve good evaluation accuracy on $\mathcal{D}_{te}$. This premise, allows us to assume a good class separation in the feature space, and a consequently well-defined set of class distributions, which will be used for constructing the references for stages \textit{(ii)} and \textit{(iii)}. We initially construct a set $\mathcal{R} = \{R(x_i), x_i, y_i\}$ of the latent representations extracted from dataset $\mathcal{D}_{tr}$ by model $f_{bsl}$, and its originating images-labels pair. The latent representations $R(x_i)$ correspond to the set of features the model originally distilled from the input allowing the FC layers to generate the class probabilities. We use such information as reference. 

The set $\mathcal{R}$ contains the latent representations for all samples of all classes in $\mathcal{D}_{tr}$. We divide $\mathcal{R}$ in $C$ sub-sets, one for each class, and we refer as $\mathcal{R}_c$. The high variability within each class images generate high variability within $\mathcal{R}_c$. Fitting such high-dimensional data, with high-variability to a distribution, would generate meaningless parameters. We address the variability by adaptively clustering semantically similar latent representations within each $\mathcal{R}_c$. We are searching for features that gravitate around a mean value and tolerate certain dispersion around this center. Given the nature of the data, we are not fixing the number of centers in the data, nor the radius of dispersion. For that, we semantically cluster our data based on the representations.

Using K-means clustering \cite{arthur2007k} we cluster the data from each $\mathcal{R}_c$. We've chosen to cluster just within each class to guarantee separation between classes. Given the set $\mathcal{R}_c$, composed of all $x_i \in \mathcal{R}_c$, we want to minimize the within cluster variance:
\begin{argmini}|s|
{\Psi}{\sum_{i=1}^{\psi} \sum_{R(x) \in \Psi_i} \mynorm{R(x)-\mu_i}_2}
{\label{eq:k-means}}{ }
\end{argmini}
where $\Psi \subset \mathcal{R}_c$, and $\psi$ is the number of cluster centers. We iteratively search for best value for $\psi$, taking into consideration that a higher value would reduce variability within clusters, but would reduce the distance between cluster centers. The objective is to find a number of clusters that balance these two factors such that the distance between samples within the same cluster is minimum and the distance between clusters is maximum.

\begin{figure*}[!t]
    \centering
    \includegraphics[width=\linewidth]{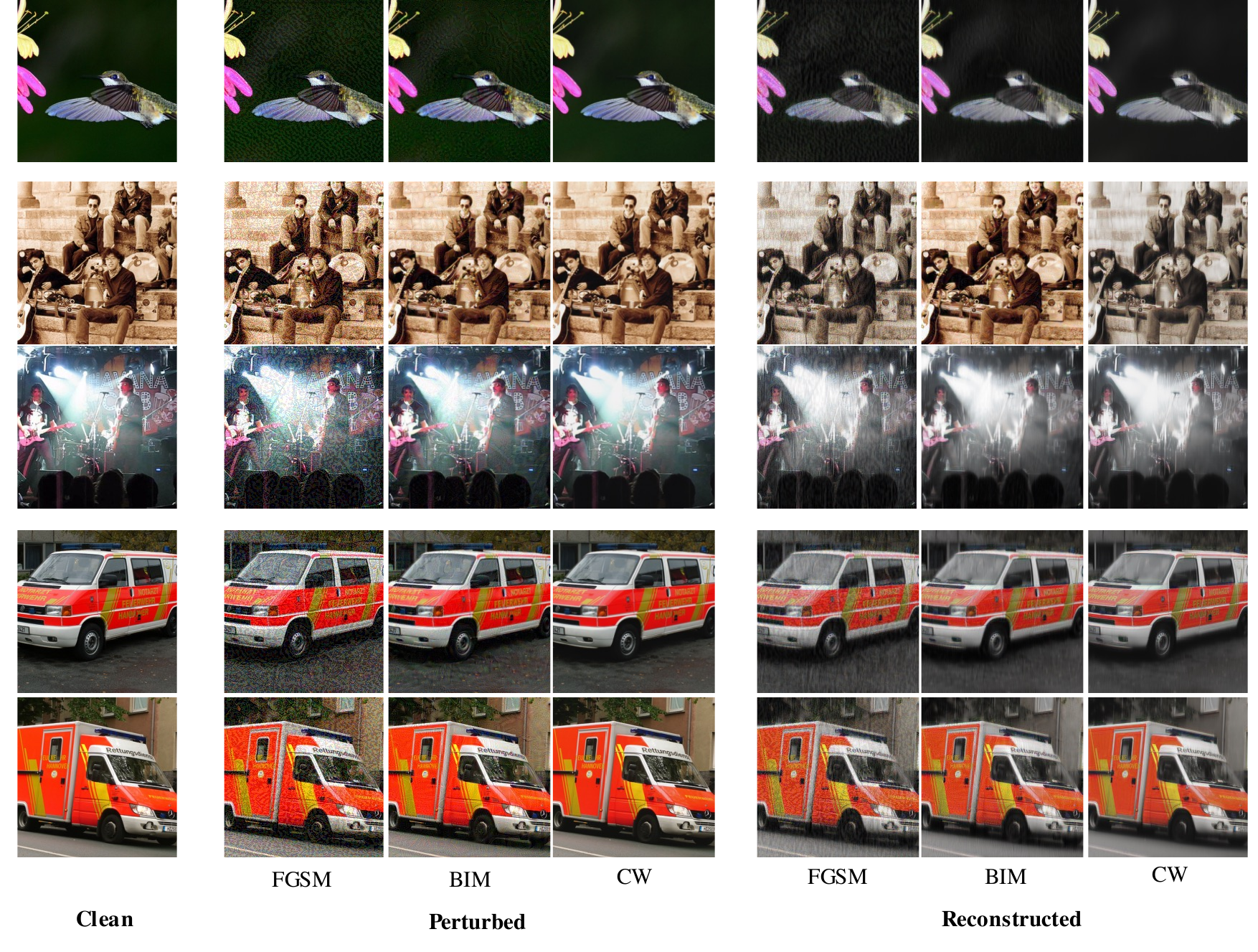}
    \caption{ The reconstruction output of RSFP on ImageNet-10 images. From left to right, first is the original clean image. The third, fourth and fifth columns shows the output of FGSM ($l_2\leq0.08$), BIM and CW ($l_2\leq0.04$). The next 3 columns show the reconstruction output of the respective attacks.}  
    \label{fig:qualitative_reconstruction}
\end{figure*}

In \autoref{eq:k-means}, the objective is to reduce the Within Cluster Sum of Squares (WCSS). We employ the elbow method to balance the cost of increasing $\psi$ with respect to the variance reduction. The WCSS is used as a performance indicator. We iterate over the value of $\psi$, smaller values on the WCSS indicates greater homogeneity within clusters, but indefinitely increasing $\psi$ will start reducing the separation of the clusters. When the the value of $\psi$ is closer to the optimal number of clusters, the WCSS curve shows a rapid decline which reduces significantly as $\psi$ increases. It is important to highlight that this process is calculated independently for each $\mathcal{R}_c$. Datasets composed of multiple classes, as seen in real applications, would not affect the performance of this algorithm. 

For each cluster $\Psi_j$, we obtain the mean $\mu_{\Psi_j} \in \mathbb{R}^k$ as the average of each individual component of each $R(x) \in \Psi_j$, and the covariance:
\begin{equation*}
    \sigma_{\Psi_j} = \mathbb{E}[(\Psi_j - \mathbb{E}[\Psi_j])(\Psi_j - \mathbb{E}[\Psi_j])^T]
\end{equation*}
where T is the transpose operator.

Each cluster represents a set of semantic features within each of the classes. These semantic features are translated from the originating images $x_i$. In stage \textit{(iii)}, we propose to reconstruct the images with dictionaries based on these clean images. We generate reconstruction dictionaries based on the known problem \textit{Convolutional Dictionary Learning} (CDL). Specifically, given a set of images $x_i \in \Psi_j$ composed of $S$ training images $\{x_t\}^S_{s=1}$, CDL is implemented through minimizing:

\begin{mini}|s|
{\{d_m\},\{r_{s,m}\}}{\frac{1}{2} \sum^S_1\mynorm{\sum^M_1 d_m*r_{s,m} -x_s}_2^2}
{\label{eq:dictionary_learning}}{ }
\breakObjective{+\lambda \sum^S_1 \sum^M_1 \mynorm{r_{s,m}}_1}
\addConstraint{\mynorm{d_m}_2 \leq 1, \forall m \in {1,...,M}}
\end{mini}
where $d_m$ are the $M$ atoms, that compose the dictionary $\Omega$, and $r_{s,m}$ are a set of coefficient maps, defined as:

\begin{mini}|s|
{\{r_m\}}{\frac{1}{2} \mynorm{\sum^M_1 d_m*r_{m} -x}_2^2 +\lambda  \sum^M_1 \mynorm{r_{m}}_1}
{\label{eq:feature_maps_learning}}{ }
\end{mini}
We observe that CDL is a computationally expensive algorithm that does not scale well to larger images and datasets. To overcome scalability issues in our method we implement the optimizations proposed by \cite{liu2018first} in an algorithmic level. Currently we use ADMM \cite{boyd2011distributed} to solve the minimization problem. The clusters, cluster distributions, and cluster reconstruction dictionaries generated for all classes are associated to create a semantic reconstruction dictionary, $\Phi = \{D, \Psi, (\mu_{\Psi}, \sigma_{\Psi})\}$.

\subsection{Robust Semantic Training}

While the adversarial attack strategy of a min-max optimization shown in \autoref{eq:robust_optimization} has shown very successful results, it fails in generalizing the method to unseen attacks. This issue is linked to the empirical solution provided to the maximization term. Since no closed form solution can be derived for such complex functions, it is often approximated by the chosen adversarial attack algorithm.  While it is very effective in adding resistance to this specific algorithm, it often fails to generalize to other attacks. We argue that the network does not learn to extract robust latent representations, but rather change the FC layers to understand latent representations extracted from adversarial samples and clean samples as being from the same class.

In our proposed solution we address this issue by introducing a constraint in the representation space. Our objective is not to change the boundaries of the decision on the FC layer, but rather extract robust semantic representations from the adversarial and clean samples that lie on the same distribution. We modify the standard adversarial training equation, adding an extra constrain in the objective function:

\begin{equation}\label{eq:semantic_robust_optimization}
\begin{split}
    \min_\theta ~ \mathbb{E}_{(x,y) \sim \mathcal{D}} \max_{\delta \leq \epsilon}&~   l(f(x'), y) +  \lambda \mynorm{\theta)}_2^2 +
    \\& \alpha(R(x') - \mu)\sigma^{-1}(R(x')-\mu))^{\frac{1}{2}}
\end{split}
\end{equation}
where the last term, the \textit{Mahalanobis Distance} (MD), minimizes the distance between the extracted adversarial latent representations $R(x')$ and the cluster distribution, following the association in $\Phi$. By minimizing the distance between the clean distribution clusters and the adversarial latent representations, the model, instead of learning the adversarial attack pattern, learns to extract meaningful representations, ignoring the added noise in the input. We refer to the robust semantic model as $f_{rob}$, and latent representations extracted from input, $x_i$, with $f_{rob}$ as $R_{rob}(x)$.

Along with our empirical evaluation, we've noticed a data-dependent constraint to our MD formulation: some clusters present covariance matrices which are not full rank. Such constraint indicates close to perfect correlation among some features in the latent representation. In this case, finding the exact inverse of the covariance matrix is not possible. To those specific cases, we've applied the Moore-Penrose pseudo-inverse \cite{barata2012moore}.

\subsection{Robust Semantic Feature Purification}

The input purification occurs only at inference time. For each input $x'_i$ we extract $R_{rob}(x'_i)$. At inference time, no label with respect to which class or cluster the input belongs to is available.
Moreover, since the input can be adversarially manipulated, it is highly important that $f_{rob}$ extracts latent representations which lie on the same distribution for both clean and adversarial images. Identifying if an input is adversarial or clean is a research question by itself and outside of the scope of this work, therefore we purify all inputs.

Based on $R_{rob}(x'_i)$, we select the semantic reconstruction dictionary which best reconstructs the high-frequency components of $x'_i$. We compute the MD between $R_{rob}(x'_i)$ and all clusters in $\Phi$. The one with minimum distance is selected as the reconstruction dictionary. In parallel, we decompose $x'_i$ into a high-frequency component, $x'_{high}$, and a low frequency component, $x'_{low}$, using the tikhonov filter \cite{garcia2018convolutional}:
\begin{argmini*}|s|
{x_{low}}{\frac{1}{2} \mynorm{x_{low}-x}_2^2 +\frac{\lambda}{2} \sum_j \mynorm{G_j x_{low}}^2_2 }
{\label{eq:tikhonov}}{ }
\end{argmini*}
where $G_j$ is an operator that computes the discrete gradient along image axis $j$. Therefore, $x'_{high} = x'-x'_{low}$. 

The reconstruction of $x'_{high}$ follows the standard sparse coding representation:
\begin{equation*}
    x_{high}^{rec} \approx Dr = d_1r_1+ \dots +d_Mr_M,
\end{equation*}
in which $D$ is the dictionary learned only from patches of clean images. Under such circumstances we have a high frequency component formed of patches of clean images, and consequently free from adversarial manipulation. The full image purification follows from adding the low and high-frequency components:
\begin{equation}
    x_{pur} = x_{low} + x_{high}^{rec}
\end{equation}

\begin{table*}[!t]
     \centering
     \caption{CIFAR-10 classification accuracy using WRN32-10 trained with PGDAT + RST.}
     \begin{tabular}{lrrrrr}
\hline
Defense & \multicolumn{1}{c}{No-attack} & \multicolumn{1}{c}{PGD} & \multicolumn{1}{c}{CW} & \multicolumn{1}{c}{BIM} & \multicolumn{1}{c}{TPGD} \\ \hline \hline
No defense & 86.36\% & 27.68\% & 9.91\% & 42.50\% & 30.37\% \\ \hline
PGDAT+RST (PGDAT) & 79.79\% & 67.43\% & 50.74\% & 62.59\% & 75.25\% \\
PGDAT (PGDAT) & 79.06\% & 66.74\% & 50.46\% & 61.25\% & 74.69\% \\
PGDAT+RST (Random) & 77.36\% & 66.99\% & 39.66\% & 60.52\% & 73.92\% \\
PGDAT (Random) & 75.63\% & 65.55\% & 52.26\% & 57.86\% & 73.57\% \\
PGDAT+RST (Trades) & 87.56\% & 73.60\% & 56.30\% & 75.27\% & 78.18\% \\ \hline \hline
\end{tabular}
    \label{tb:adversarial_training}
\end{table*}

\begin{table*}[!t]
    \centering
    \caption{CIFAR-10 classification accuracy against adversarial across different models when the input is purified with RSFP.}
    \begin{tabular}{l|r|rrrrr}
\hline
Defense  & \multicolumn{1}{c}{No-attack} & \multicolumn{1}{c}{FGSM-0.08} & \multicolumn{1}{c}{FGSM-0.04} & \multicolumn{1}{c}{BIM} & \multicolumn{1}{c}{DeepFool} & \multicolumn{1}{c}{CW} \\ \hline \hline
AlexNet &  91.07\% & 73.07\% & 76.78\% & 83.92\% & 82.69\% & 86.53\% \\ 
VGG-16 &  94.23\% & 78.57\% & 75.00\% & 76.78\% & 83.92\% & 87.05\% \\
ResNet50 &  95.19\% & 76.78\% & 86.53\% & 84.61\% & 90.38\% & 90.78\% \\
GoogleNet  & 90.38\% & 79.80\% & 83.65\% & 87.5\% & 88.5\% & 85.57\% \\ \hline \hline
\end{tabular}
    \label{tb:multiple_models}
\end{table*}

\begin{table*}[!t]
    \centering
    \caption{CIFAR-10 classification accuracy using VGG-16 on images reconstructed with RSFP. 'No Defense' indicates no image reconstruction was applied.}
    
\begin{tabular}{l|r|rrrrr}

\hline
\multicolumn{1}{c|}{\textbf{Defense}} & \multicolumn{1}{c|}{\textbf{Clean}} & \multicolumn{1}{c}{\textbf{FGSM-0.08}} & \multicolumn{1}{c}{\textbf{FGSM-0.04}} & \multicolumn{1}{c}{\textbf{BIM}} & \multicolumn{1}{c}{\textbf{DeepFool}} & \multicolumn{1}{c}{\textbf{CW}} \\ \hline \hline
No defense & 94.23 & 58.16 & 65.23 & 18.03 & 17.60 & 9.36 \\ \hline
MagNet  & 90.35 & 61.45 & 65.21 & 43.12 & 65.35 & 48.45 \\
PixelDefend  & 85.26 & 68.10 & 73.29 & 77.29 & 74.14 & 75.79 \\
STL & 83.60 & 71.03 & 75.47 & 75.31 & 79.59 & 79.06 \\ \hline
RSFP  & 94.23 & 78.57 & 75.00 & 76.78 & 83.92 & 87.50 \\ \hline \hline
\end{tabular}

    \label{cifar_table1}
\end{table*}

\begin{table*}[!t]
    \centering
   
    \caption{ImageNet-10 classification accuracy using VGG-16. In images with resolution 64x64 and resolution 128x128.}
    
    \begin{tabular}{lllllll}
\hline
\multicolumn{7}{c}{\textbf{Resolution 64x64}} \\ \hline
\multicolumn{1}{l|}{{\color[HTML]{000000} Defense}} & \multicolumn{1}{l|}{{\color[HTML]{000000} Clean}} & {\color[HTML]{000000} FGSM-0.08} & {\color[HTML]{000000} FGSM-0.04} & {\color[HTML]{000000} BIM} & {\color[HTML]{000000} DeepFool} & {\color[HTML]{000000} CW} \\ \hline \hline
\multicolumn{1}{l|}{No Defense} & \multicolumn{1}{l|}{86.65} & 28.16 & 30.8 & 18.83 & 8.11 & 7.51 \\ \hline
\multicolumn{1}{l|}{TVM} & \multicolumn{1}{l|}{75.55} & 59.97 & 69.3 & 71.56 & 72.1 & 71.87 \\
\multicolumn{1}{l|}{Quilting} & \multicolumn{1}{l|}{77.41} & 73.04 & 74.18 & 76.42 & 76.46 & 76.62 \\
\multicolumn{1}{l|}{Crop-Ens} & \multicolumn{1}{l|}{75.08} & 69.68 & 72.21 & 73.69 & 74.01 & 73.04 \\
\multicolumn{1}{l|}{PD-Ens} & \multicolumn{1}{l|}{82.5} & 66.34 & 76.07 & 79.03 & 79.55 & 78.13 \\
\multicolumn{1}{l|}{STL} & \multicolumn{1}{l|}{84.21} & 75.14 & 80.38 & 81.03 & 82.21 & 81.22 \\ \hline
RSFP &87.50& 84.37 & 78.12 & 87.50 & 84.37 & 81.25 \\ \hline \hline
\multicolumn{7}{c}{\textbf{Resolution 128x128}} \\ \hline
\hline
\multicolumn{1}{l|}{{\color[HTML]{000000} Defense}} & \multicolumn{1}{l|}{{\color[HTML]{000000} Clean}} & {\color[HTML]{000000} FGSM-0.08} & {\color[HTML]{000000} FGSM-0.04} & {\color[HTML]{000000} BIM} & {\color[HTML]{000000} DeepFool} & {\color[HTML]{000000} CW} \\ \hline \hline
\multicolumn{1}{l|}{No Defense} & \multicolumn{1}{l|}{89.91} & 21.23 & 24.09 & 17.90 & 5.84 & 5.04 \\ \hline
\multicolumn{1}{l|}{TVM} & \multicolumn{1}{l|}{85.91} & 25.68 & 43.86 & 65.86 & 63.60 & 61.29 \\
\multicolumn{1}{l|}{Quilting} & \multicolumn{1}{l|}{81.49} & 39.03 & 58.89 & 64.34 & 62.42 & 59.22 \\
\multicolumn{1}{l|}{Crop-Ens} & \multicolumn{1}{l|}{77.30} & 46.22 & 64.47 & 68.76 & 70.60 & 68.88 \\
\multicolumn{1}{l|}{PD-Ens} & \multicolumn{1}{l|}{87.89} & 23.33 & 42.86 & 72.21 & 73.59 & 72.72 \\
\multicolumn{1}{l|}{STL} & \multicolumn{1}{l|}{86.54} & 47.33 & 66.06 & 73.23 & 73.01 & 74.32 \\ \hline
RSFP &87.50& 87.25 & 89.28 & 87.5 & 90.62 & 88.25 \\ \hline \hline
\end{tabular}
     \label{imagenet_vgg_table1-a}
\end{table*}

\section{Experiments}
\subsection{Experimental Settings}
We evaluate our model on 2 main datasets, CIFAR-10 \cite{krizhevsky2009learning}, and ImageNet \cite{ILSVRC15}, from which we extract 10 classes to compose ImageNet-10. CIFAR-10 is composed of 60000 images (50k for training and 10k for testing), uniformly distributed among 10 classes. ImageNet is composed of 1000 classes with roughly 1300 images per class for training and 50 samples for testing per class. 

To evaluate the efficiency of our model, we use VGG16 \cite{simonyan2014very} and ResNet-50 \cite{he2016deep} as classification models for comparison with other defense methods. We attack our model with FGSM\cite{goodfellow2015explaining}, BIM \cite{kurakin2017adversarial}, DeepFool \cite{moosavi2016deepfool}, and CW \cite{carlini2017towards}. For FGSM, we evaluate under the $l_2 norm <0.04$ and $l_2 norm <0.08$, and for BIM, DeepFool and CW we restrict to $l_2 norm <0.04$ and 100 iteration steps. The images in the dataset are normalized in the range between 0 and 1. For all experiments a single model is robustly trained to extract robust features for each dataset evaluated.

We provide a qualitative evaluation of the effects of dictionary selection based on the semantic cluster. Even though PSNR can measure the signal to noise ratio, we've observed it is not a good metric for adversarial comparison, as attacks like CW and DeepFool are efficient in adding perturbations without changing PSNR significantly. In all the tables, we refer to our image purification as Robust Semantic Feature Purification (RSFP), and the Robust Semantic Training as RST.

\subsection{Robust Semantic Training}

We first evaluate the effectiveness of adversarial training using the adaptive semantic clusters as a reference for the cluster approximation and the generalization capabilities of our method. For that, we use CIFAR-10 dataset, and a WideResNet32-10 (WRN32-10), as proposed in \cite{song2019robust}. Moreover, we adversarially train WRN32-10 using the standard adversarial training method approximating the maximization with Projected Gradient Descend attack \cite{Madry2018}, in which 10 steps are used, with $l_2$ norm perturbation limit of $\delta = 0.3$.

\autoref{tb:adversarial_training} shows the accuracy of our model minimizing the distance between the distribution of reference clusters and the model's extracted latent representation. We've used the PGD adversarial training in composition with the Robust Semantic Training (RST). In parenthesis we've indicated if we used any pre-trained weights set, as initialization for our model parameters, (Random) indicates no pre-training.

As seen in \autoref{tb:adversarial_training}, we have evaluated our model under several conditions, including transfer learning from other techniques (We've restricted to only using other techniques as weight initialization, no modifications to \autoref{eq:semantic_robust_optimization} was used). 
We've observed that when combined with transfer learning from other techniques such as TRADES \cite{Zhang2018} we achieve good performance, but not when training from initial random weights. 
It is important to highlight that this mechanism is meant only to extract robust latent representations, it is not meant to be a standalone defense.

\subsection{Robust Semantic Feature Purification Evaluation}

An advantage of our defense method is the transferability among multiple attacks and across multiple models. \autoref{fig:qualitative_reconstruction} shows the reconstruction capabilities of our model under different attacks for samples in ImageNet-10. \autoref{tb:multiple_models} shows a quantitative evaluation of our model to defend AlexNet, VGG16, GoogleNet, and ResNet50 attacked by FGSM, BIM, DeepFool, and CW. We train a single model $f_{rob}$ to extract the robust latent representations and defend any of the models in \autoref{tb:multiple_models}.

\begin{figure}[!b]
    \centering
    \includegraphics[width=\linewidth]{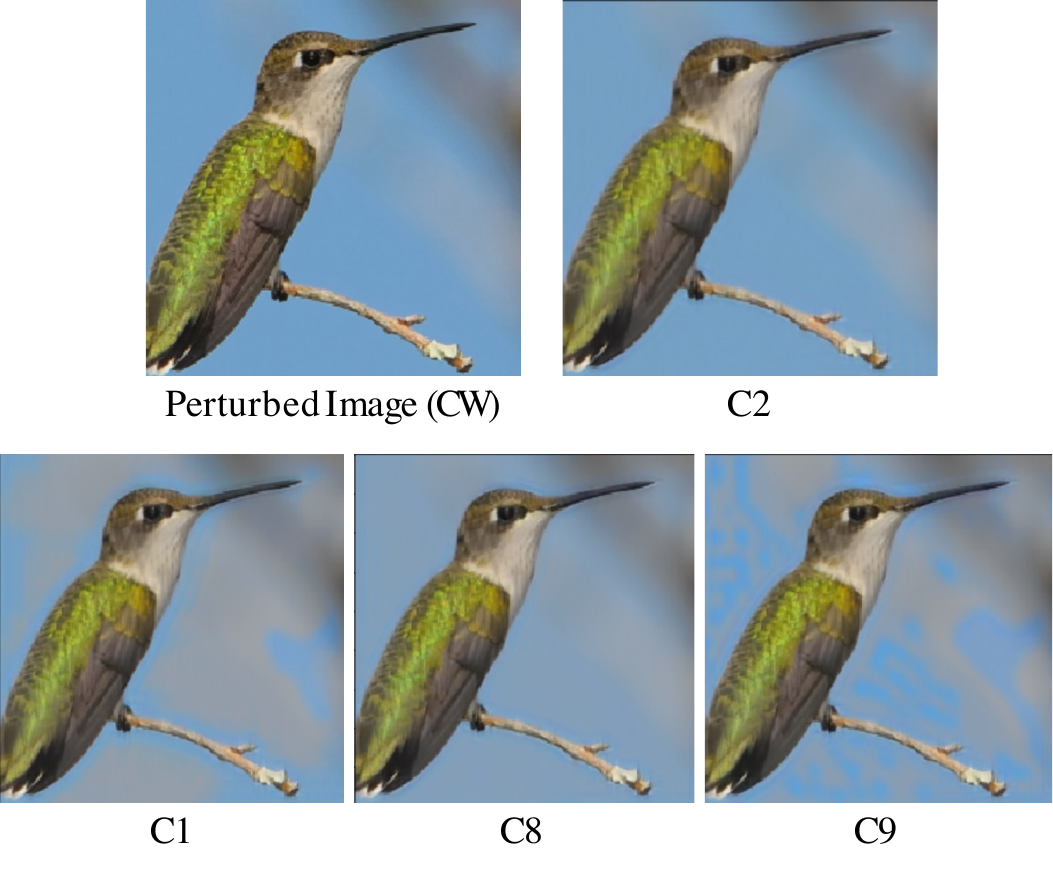}
    \caption{ Qualitative analysis of the reconstruction based on the cluster selection. Top images are attacked and best reconstruction. C1, C8, and C9 are reconstructions from manually selected clusters.}
    \label{fig:cluster_influence}
\end{figure}

We compare our model to other robust defenses that involve input transformation. In Table \ref{cifar_table1}, we use a pre-trained VGG-16, adjusting the parameters and output layer for CIFAR-10, and obtain the accuracy against the mentioned attacks and compare against defenses such as MagNet\cite{meng2017magnet}, PixelDefend \cite{song2018pixeldefend}, and STL \cite{sun2019adversarial}, following the experimental setting of \cite{sun2019adversarial}. 

\autoref{imagenet_vgg_table1-a} summarizes our results against the defenses proposed in \cite{guo2018countering} (TVM, Quilting, Crop-Ens), \cite {prakash2018deflecting} (PD-Ens) and \cite{sun2019adversarial} (STL), in ImageNet-10 for image resolutions of 64x64 and 128x128.\\

\textbf{Ablation Studies:} We've evaluated the efficiency of our model with and without the influence of transfer learning. As seen on \autoref{tb:adversarial_training}, our adversarial model achieves better results when trained with already saturated models, providing an improvement in the accuracy. Consequently, we improve the clean data accuracy which is lost on standard adversarial training. 

Moreover, we've evaluated the effects of using our current setting, pre-trained (ImageNet) classification models, against the ones trained from scratch. For CIFAR-10, clean images attacked with CW, classified with vanilla VGG-16 trained from scratch, an accuracy of 61.53\% was obtained, in contrast to our reported 87.5\%. This shows that transfer learning also contributes to our defense. Without our defense, in both cases, the models performed poorly, achieving only 9.36\% accuracy for the transfer learning model. Qualitative evaluation reveals that the reconstruction quality is dependent on the cluster selection, as illustrated in \autoref{fig:cluster_influence}.

\section{Discussion}

Comparison results show that our method outperforms current state-of-the-art input transformation methods based on image transformation and on sparse code image reconstruction at defending models against gray-box attacks. We assume the attacker has full knowledge of the model, but no awareness of the transformation itself.

We’ve shown that our defense is model agnostic and maintains accuracy across different models and several unseen attacks. We have associated this fact to both our robust models ability to approximate the robust latent representations to the clean distribution, and the fact that we learn sparse dictionaries from clean images and use them for the reconstructions, which makes our models minimally dependent on the adversarial attack's empirical approximation of \autoref{eq:robust_optimization}. 

\begin{figure}[!t]
    \centering
    \includegraphics[width=\linewidth]{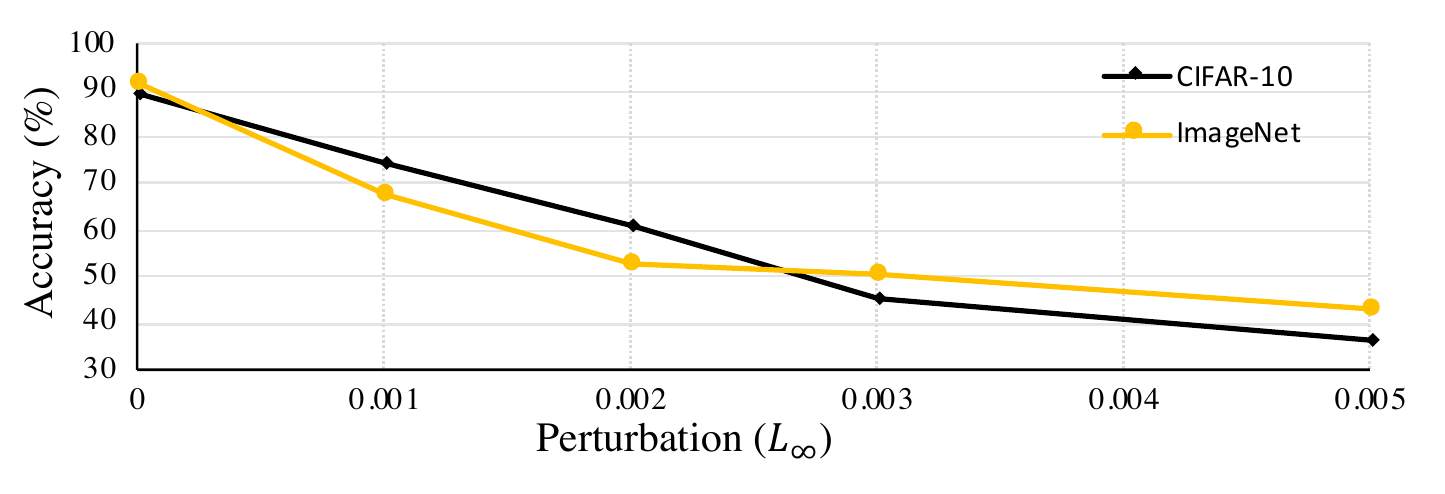}
    \caption{ Accuracy evaluation on CIFAR-10 (yellow miters) of Resnet50  and VGG-16 on ImageNet-10 FGSM under $l_\infty$ norm.}
    \label{fig:tradeoff_linf}
\end{figure}

The assumption of the class separation based on the accuracy has proven reasonable withing the set of experiments proposed in this manuscript. With the increase in the number of classes it would be expected that a reduction in the distance between the distributions would occur, leading to a reduction of the accuracy of the algorithm. But in a high-dimensional setting, common in the SOTA models used for this vision task, $R(x) \in \mathbb{R}^d$ generally has a high-dimensional feature space ($d \geq 2048$). With such high-dimensional feature space we can enforce the class separation through the training loss, hence the effect of increasing the number of classes is almost indifferent for the overall accuracy. Moreover, the increase of the number of classes as well as the resolution of the images, as discussed in \autoref{baseline_training}, could lead \autoref{eq:feature_maps_learning} to be unsolvable due to the exponential computational cost. But our implementation of the CBPDN problem, based on the optimizations proposed in \cite{liu2018first}, allowed a scalable solution to this problem. 

All the experiments we've provided demonstrate the efficiency of our method to attacks in which the space norm is defined by $l_2$ norm. We've observed that $l_2$ bounded attacks generate less visible corruption on the input. We've also evaluated our model under $l_\infty$ norm bounded FGSM. In \autoref{fig:tradeoff_linf} we show the efficiency of our defense method under different levels of $l_\infty$ attack. As we see, as the perturbation level increases, it starts affecting the low-frequency components of the image, and our method cannot purify these images. Therefore methods involving GAN's and Autoencoders for the full reconstruction are necessary.

\section{Conclusion}

We have proposed a novel adaptive Robust Semantic Feature Purification defense that presents SOTA results against $l_2$ bounded adversarial attacks, unseen at training time. We design a new methodology for input transformation which creates semantic reconstruction dictionaries for high-frequency components for each cluster of latent representations of the images in our dataset. We evaluate our proposed methodology on CIFAR-10 and ImageNet, and we've shown that our defense method achieves robustness against several unseen attacks and different target models (AlexNet, VGG-16, GoogleNet, and ResNet50). We have also evaluated our model against $l_\infty$ bounded perturbations and have seen a less effective transformation. These qualitative and quantitative results on $l_\infty$ perturbations indicate that when the corruption achieves lower frequency portions of the image, the image needs to be regenerated rather than purified, which indicates the need of generative approaches.


\bibliography{example_paper}
\bibliographystyle{IEEEtran}

\begin{IEEEbiography}
[{\includegraphics[width=1in,height=1.3in,clip,keepaspectratio]{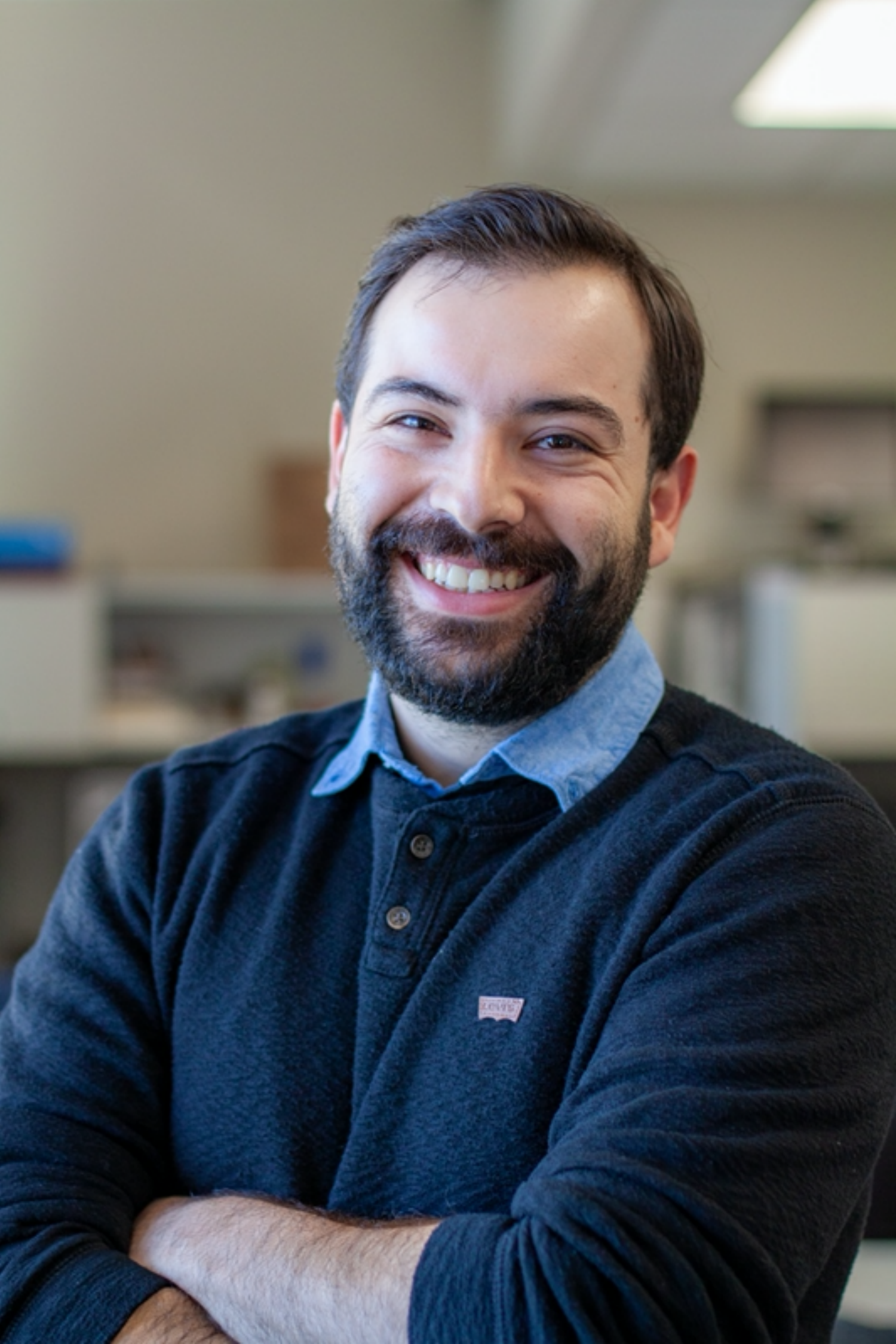}}]{Samuel Henrique Silva}
is currently a Ph.D. student and research fellow at the Open Cloud Institute of University of Texas at San Antonio (UTSA), San Antonio, TX, USA. Samuel received the Bachelor of Science (B.Sc.) degree in Control and Automation Engineering from State University of Campinas, Campinas, Brazil, in 2012 and the M.S. degree in Electrical Engineering from the University of Notre Dame, Notre Dame, IN, USA in 2016. He is a member of the IEEE, Eta Kappa Nu honor society. Samuel's research interests are in the areas of artificial intelligence, robustness in deep learning models, autonomous decision making, multi-agent systems and adversarial environments.
\end{IEEEbiography}

\begin{IEEEbiography}
[{\includegraphics[width=1in,height=1.25in,clip,keepaspectratio]{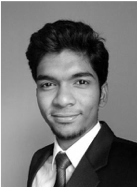}}]{Arun Das} is currently a Ph.D. student and research fellow at the Open Cloud Institute of University of Texas at San Antonio (UTSA), San Antonio, TX, USA. Arun received the Bachelor of Technology (B.Tech.) degree in Electrical and Electronics Engineering from Cochin University of Science and Technology, Kerala, India, in 2013 and the M.S. degree in Computer Engineering from the University of Texas at San Antonio, San Antonio, TX, USA in 2016. He is a member of the IEEE, and IEEE Eta Kappa Nu honor society. Arun's research interests are in the areas of artificial intelligence, computer vision, distributed and parallel computing, cloud computing and computer architecture.
\end{IEEEbiography}

\begin{IEEEbiography}
[{\includegraphics[width=1in,height=1.25in,clip,keepaspectratio]{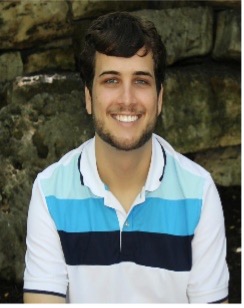}}]{Ian Scarff} is a PhD student and GRA at the Secure AI and Autonomy Lab at the University of Texas at San Antonio (UTSA). He received his Bachelors of Science in Statistics from Texas A\&M University in 2015 and his Masters of Science in Data Analytics in 2020 from UTSA. He currently studies Information Technology. His interests are in artificial intelligence, computer vision, and autonomy.
\end{IEEEbiography}

\begin{IEEEbiography}
[{\includegraphics[width=1in,height=1.3in,clip,keepaspectratio]{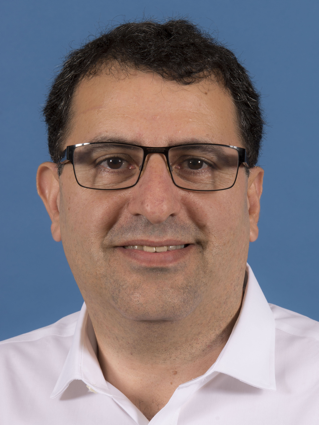}}]{Peyman Najafirad}
is a co-founder and Associate Director of the Open Cloud Institute (OCI), and an Associate Professor with the Information Systems and Cyber Security Department at the University of Texas at San Antonio. He received his first B.S. degree from Sharif University of Technology in Computer Engineering in 1994, his 1st master in artificial intelligence from the Tehran Polytechnic, his 2nd master in computer science from the University of Texas at San Antonio (Magna Cum Laude) in 1999, and his Ph.D. in electrical and computer engineering from the University of Texas at San Antonio. He was a recipient of the Most Outstanding Graduate Student in the College of Engineering, 2016, earned the Rackspace Innovation Mentor Program Award for establishing Rackspace patent community board structure and mentoring employees (2012), earned the Dell Corporation Company Excellence (ACE) Award  for exceptional performance and innovative product research and development contributions (2007), and earned the Dell Inventor Milestone Award, Top 3 Dell Inventor of the year (2005). He holds 15 U.S. patents on cyber infrastructure, cloud computing, and big data analytics with over 300 product citations by top fortune 500 leading technology companies such as Amazon, Microsoft, IBM, Cisco, Amazon Technologies, HP, and VMware. He has advised over 200 companies on cloud computing and data analytics with over 50 keynote presentations. High performance cloud group chair at the Cloud Advisory Council (CAC), OpenStack Foundation Member (the \#1 open source cloud software), San Antonio Tech Bloc Founding Member, and Children’s Hospital of San Antonio Foundation board member.
\end{IEEEbiography}

\vfill

\end{document}